\documentclass[conference]{IEEEtran}
\IEEEoverridecommandlockouts
\usepackage{cite}
\usepackage{amsmath,amssymb,amsfonts}
\usepackage{algorithmic}
\usepackage{graphicx}
\usepackage{textcomp}
\usepackage{xcolor}
\usepackage{amsmath}
\usepackage{graphicx}
\usepackage{multirow}
\usepackage{url}
\usepackage{hyperref}
\def\BibTeX{{\rm B\kern-.05em{\sc i\kern-.025em b}\kern-.08em
    T\kern-.1667em\lower.7ex\hbox{E}\kern-.125emX}}
\begin{document}

\title{DLEN: Dual Branch of Transformer for Low-Light Image Enhancement in Dual Domains}

 \author{
 \IEEEauthorblockN{Junyu Xia$^{1}$\IEEEauthorrefmark{1}}
 \IEEEauthorblockA{
 \textit{Shanghai University}\\
 lalalo@shu.edu.cn}
 \and
 \IEEEauthorblockN{Jiesong Bai$^{2}$}
 \IEEEauthorblockA{
 \textit{Shanghai University}\\
 jiesongbai.7@gmail.com}
 \and
 \IEEEauthorblockN{Yihang Dong$^{3}$}
 \IEEEauthorblockA{
 \textit{University of Chinese Academy of Sciences}\\
 dongyihang23@mails.ucas.ac.cn}
 \thanks{\IEEEauthorrefmark{1}Junyu Xia is the corresponding author.}
 }

\maketitle

\begin{abstract}
Low-light image enhancement (LLE) aims to improve the visual quality of images captured in poorly lit conditions, which often suffer from low brightness, low contrast, noise, and color distortions. These issues hinder the performance of computer vision tasks such as object detection, facial recognition, and autonomous driving. Traditional enhancement techniques, such as multi-scale fusion and histogram equalization, fail to preserve fine details and often struggle with maintaining the natural appearance of enhanced images under complex lighting conditions. Although the Retinex theory provides a foundation for image decomposition, it often amplifies noise, leading to suboptimal image quality. In this paper, we propose the Dual Light Enhance Network (DLEN), a novel architecture that incorporates two distinct attention mechanisms, considering both spatial and frequency domains. Our model introduces a learnable wavelet transform module in the illumination estimation phase, preserving high- and low-frequency components to enhance edge and texture details. Additionally, we design a dual-branch structure that leverages the power of the Transformer architecture to enhance both the illumination and structural components of the image. Through extensive experiments, our model outperforms state-of-the-art methods on standard benchmarks. The code is available at:\url{https://github.com/LaLaLoXX/DLEN}
\end{abstract}


\section{Introduction}
Low-light image enhancement (LLE) constitutes a pivotal area of study within the broader fields of computer vision and image processing. This task is fundamentally aimed at improving visual quality and perceptual clarity of images captured in suboptimal lighting conditions, thereby enhancing their interpretability and functional utility. Images acquired under low-light environments often suffer from insufficient brightness, low contrast, increased noise, and pronounced color distortions. These limitations can critically undermine performance in various downstream computer vision applications, such as object detection, facial recognition, and autonomous driving systems. Therefore, LLE has become an essential area of research in computer vision and image processing.

Traditional image enhancement techniques, such as multi-scale fusion \cite{ref56}, \cite{ref57}, homomorphic filtering \cite{ref58},  \cite{ref59}, and histogram equalization \cite{ref1}, \cite{ref62}, have been widely explored. However, these methods often rely on relatively simple algorithms, and they suffer from significant limitations, particularly in preserving fine details. They frequently fail to maintain the natural appearance of enhanced images, especially under complex and varying lighting conditions. The classical Retinex theory \cite{ref7} introduced a novel framework for image decomposition, suggesting that a color image can be separated into two components: illumination and reflection. Although this approach often amplifies noise and results in suboptimal image quality, it has provided a foundational basis for addressing modern challenges in low-light image enhancement.

\begin{figure}[ht]
\centering
\includegraphics[width=0.9\columnwidth]{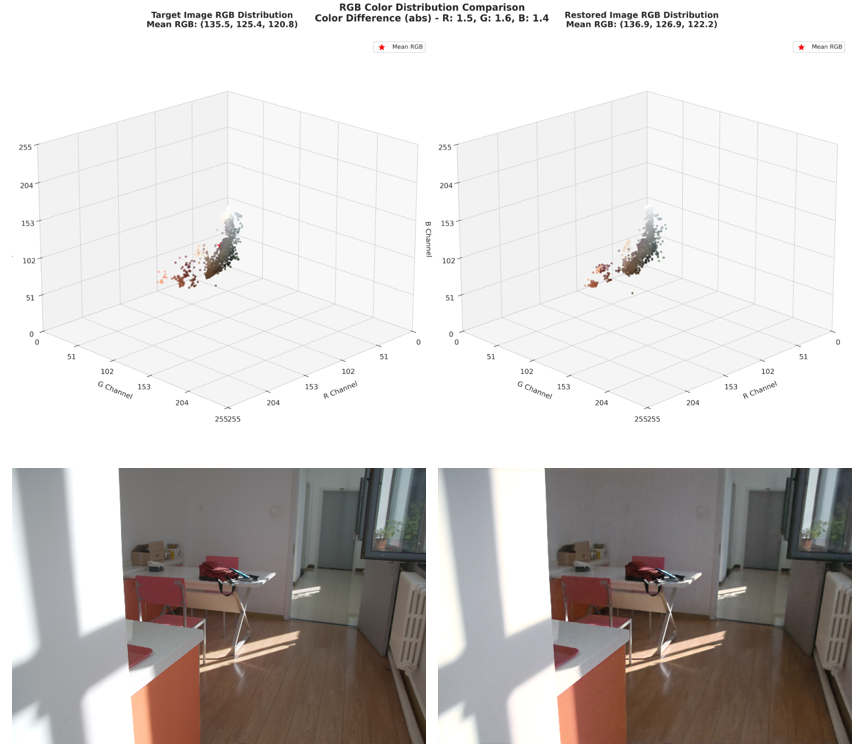} 
\caption{The target image (left) and the enhanced result (right) produced by our method. Our approach effectively enhances the image in color space.}
\label{fig1}
\vspace{-10pt}
\end{figure}

\begin{figure*}[ht]
\centering
\includegraphics[width=\textwidth]{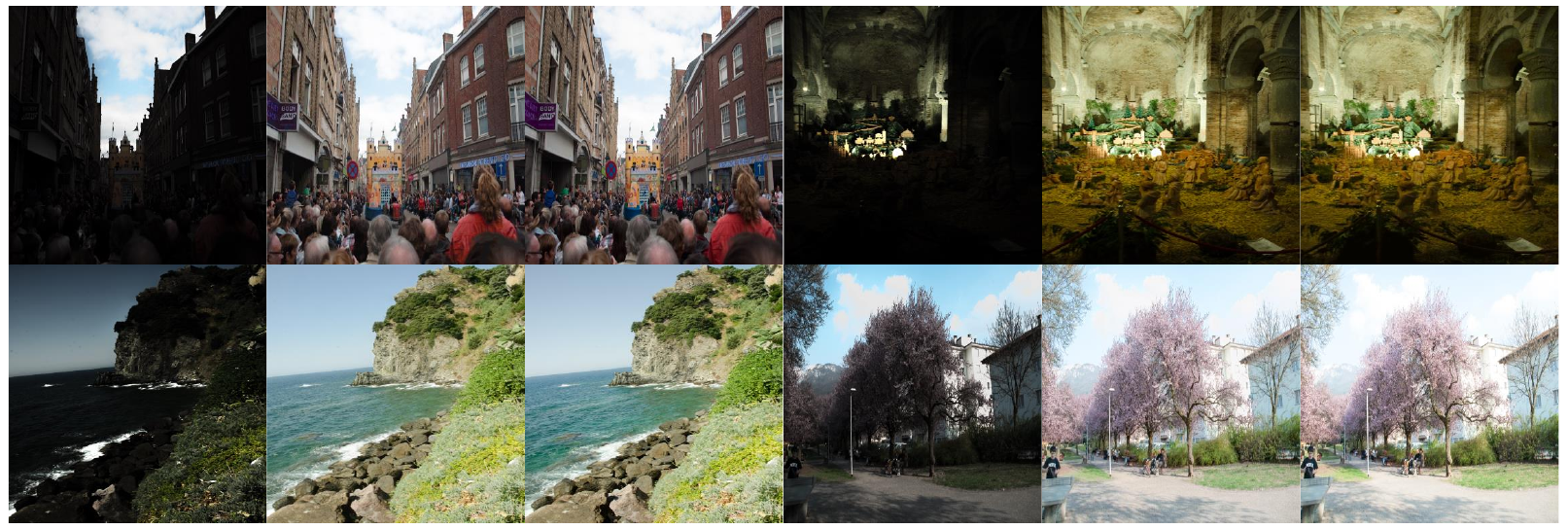} 
\caption{Visualized results on a benchmark dataset. Each image shows the input image, the output produced by our method, and the corresponding target image.}
\label{fig2}
\vspace{-10pt}
\end{figure*}

Recent advancements in neural networks, particularly convolutional neural networks (CNNs), transformer models, and mamba \cite{ref61}, have set new benchmarks in low-light image enhancement. CNNs \cite{ref8}, \cite{ref60} are effective and adaptable, leveraging local feature extraction and pooling. However, they are limited by their reliance on local receptive fields, which restricts their ability to capture global illumination. Additionally, their dependence on large labeled datasets poses challenges for generalizing across diverse low-light scenarios. In contrast, transformer models \cite{ref11}, \cite{ref14}, with their multi-head self-attention mechanism, excel at capturing global illumination and modeling long-range pixel relationships. Incorporating illumination-aware modules has further improved their ability to produce natural enhancement results. However, transformers may introduce artifacts due to over-enhancement and are highly dependent on domain-specific data, making it difficult to train a universal model for a wide variety of low-light conditions, such as nighttime, indoor, or foggy scenes.

In this paper, we propose the \textbf{Dual Light Enhance Network (DLEN)}, a novel architecture for low-light image enhancement that incorporates two distinct attention mechanisms, simultaneously considering both the spatial and frequency domains. Our network structure builds upon the work in \cite{ref11} and consists of two main components: the illumination estimation module and the restorer module. Unlike \cite{ref11}, we introduce a learnable wavelet transform module in the illumination estimation phase. This module effectively learns and preserves both high-frequency and low-frequency components, significantly enhancing edge and texture details under extremely low-light conditions. Furthermore, we design a novel dual-branch restorer structure that leverages the expressive power of the Transformer architecture. The first branch (MIAB) enhances different regions of the image based on illumination guidance, while the second branch (SEAB) focuses on enhancing and repairing the structural information of the image, with both branches being fused for the final output.

Through extensive experiments, we present both quantitative and qualitative results that highlight the superiority of our model on standard benchmarks. As demonstrated in Tab. \ref{tab:mytable1}, our approach outperforms state-of-the-art deep learning methods on the LOL dataset \cite{ref19} \cite{ref20}. 
\vspace{-10pt}
\subsection*{Our contributions are summarized as follows:}
\begin{enumerate}
    \item We introduce a learnable wavelet transform module that captures and restores high-frequency details during the enhancement process, enabling our model to perform enhancement in both the spatial and frequency domains.
    \item We design a dual-branch enhancement structure that effectively leverages the expressive power of the Transformer architecture. This structure facilitates spatial interactions and learning across various regions under different conditions, significantly improving both the illumination and structural components of the image.
    \item Extensive qualitative and quantitative experiments demonstrate that our model outperforms current mainstream baseline methods.
\end{enumerate}

\section{Related work}

\subsection{Low-Light Image Enhancement}
\textbf{Direct Enhancement Methods}:
Traditional methods, such as Histogram Equalization \cite{ref4} and Gamma Correction  \cite{ref23} , improve contrast and brightness by adjusting the image's gray-level distribution or by applying nonlinear functions to modify pixel values. These methods are simple and effective, but they may lead to over-enhancement or loss of details, especially in high dynamic range scenes. Moreover, these methods are sensitive to parameter selection, causing color distortion issues.

\textbf{Traditional Enhancement Methods}:
Traditional methods decompose an image into illumination and reflection components \cite{ref7}, simulating human visual perception of brightness and color to significantly enhance the representation of contrast and detail. For example, the models proposed by \cite{ref63}, \cite{ref66}  have shown excellent performance in applications such as underwater imaging, remote sensing, and foggy or dusty environments. However, these methods \cite{ref26}, \cite{ref27} often struggle to handle noise effectively and exhibit poor performance in terms of detail preservation. 

\textbf{Deep Learning Methods}:
Deep learning methods for low-light image enhancement have led the field since \cite{ref33}. Following the Retinex theory \cite{ref7}, CNN-based approaches \cite{ref34}, \cite{ref40} became widely adopted, with \cite{ref30}, \cite{ref40} combining Retinex decomposition with deep learning. However, these methods are struggled with long-range dependencies and \cite{ref11} addressed this by introducing transformer architectures, though these models face computational challenges with long sequences.

\subsection{Low Level Vision Transformer}
The Transformer model, originally proposed by \cite{ref12}, is a neural network architecture designed for natural language processing. It captures global dependencies by computing pairwise interactions between input elements through a self-attention mechanism \cite{ref67} \cite{ref68}. In recent years, Transformers have been applied to low-light image enhancement \cite{ref14} \cite{ref16}, addressing the limitations of CNNs in modeling long-range dependencies. For instance, \cite{ref41} introduced the SNR-Net model, which integrates global Transformer layers into a U-shaped CNN to enhance structural understanding. \cite{ref11} combines Retinex theory with a single-stage Transformer framework, achieving better noise reduction and structural consistency. Butthe high computational cost and low interpretability of Transformer models remain significant obstacles to their application in real-time scenarios.

\subsection{Wavelet Transforms Application}
Wavelet transforms have become a critical tool in addressing low-level vision tasks by leveraging frequency-domain analysis. In image denoising, wavelet-based methods  \cite{ref53} \cite{ref54} effectively separate high-frequency noise from low-frequency signals, enhancing restoration quality. Similarly, wavelet techniques have been applied to image super-resolution \cite{ref60}, focusing on high-frequency components to recover fine details. In image de-blurring, wavelet transforms enable efficient modeling of motion blur trajectories, outperforming traditional spatial-domain methods. Recent advancements include \cite{ref55}, a multiscale motion-deblurring network that integrates a learnable discrete wavelet transform module. By capturing directional continuity and frequency features, \cite{ref55} simplifies coarse-to-fine schemes and achieves state-of-the-art performance across multiple datasets.

\begin{figure*}[ht]
\centering
\includegraphics[width=\textwidth]{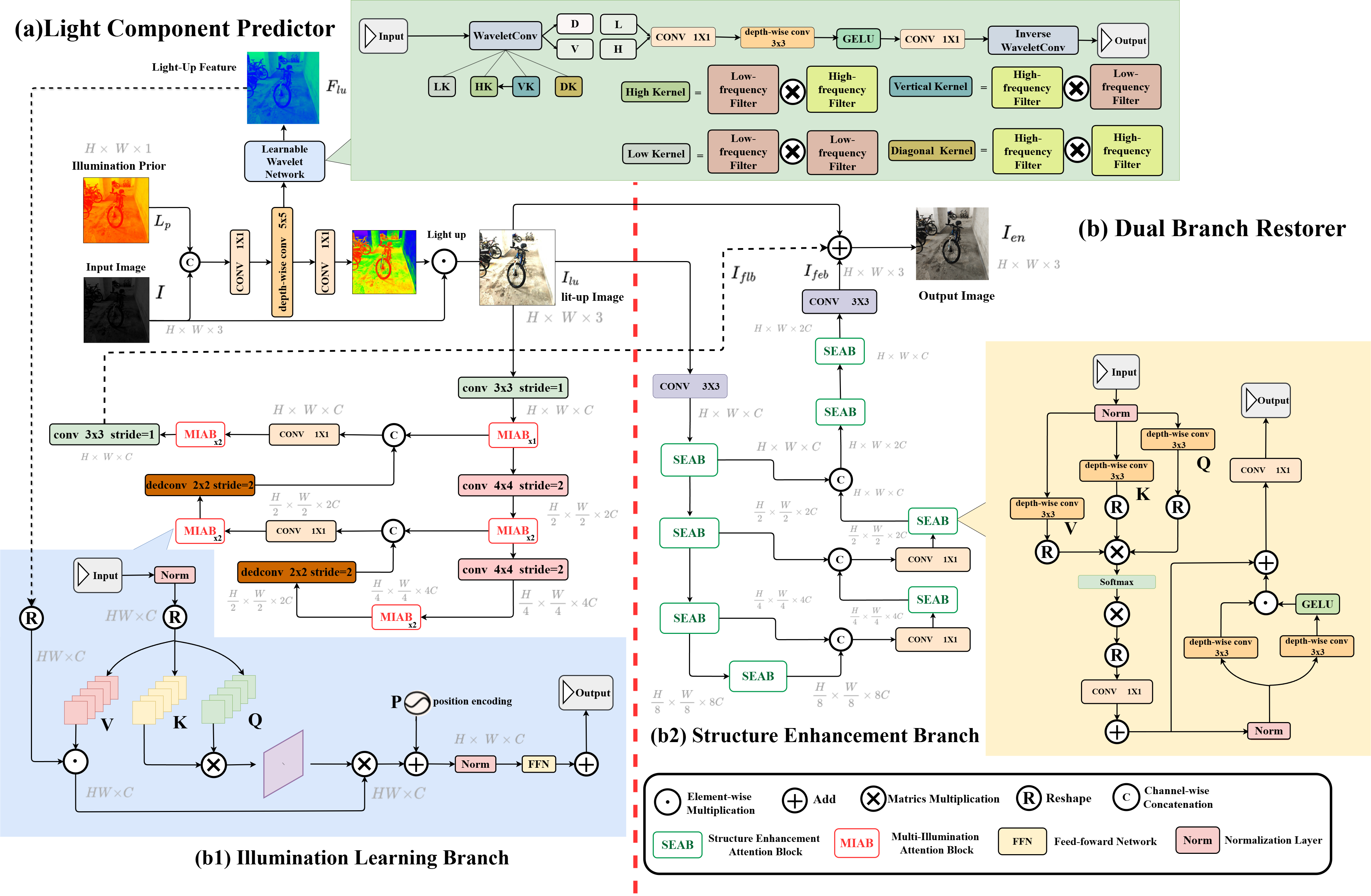} 
\caption{The figure illustrates the detailed structure of our model, which consists of two main components: (a) the Light Component Predictor and (b) the Dual-Branch Restorer.
}
\label{fig6}
\vspace{-10pt}
\end{figure*}
\section{Methods}
As shown in Figure.\ref{fig6}, our architecture comprises two primary components: the Light Component Predictor (LCP) and the Dual-Branch Restorer (DBR). Drawing inspiration from the Retinex theory \cite{ref7}, the LCP consists of multiple convolutional layers, followed by a learnable wavelet transform network designed to capture various illumination effects. The DBR, on the other hand, is built upon two essential units: the Multi-Illumination Attention Block (MIAB) and the Structure Enhancement Attention Block (SEAB), as depicted in Figure.\ref{fig6}(b). These components work in tandem to improve restoration quality under varying lighting conditions.

\subsection{Overall Pipeline} According to the Retinex theory, a low-light image $\mathbf{I} \in \mathbb{R}^{H \times W \times 3}$ can be decomposed into a reflection image $\mathbf{R} \in \mathbb{R}^{H \times W \times 3}$ and an illumination image $\mathbf{L} \in \mathbb{R}^{H \times W}$, expressed as: 
\vspace{-4pt}
\begin{equation} \mathbf{I} = \mathbf{R} \odot \mathbf{L}, \end{equation} 
where $\odot$ denotes element-wise multiplication. The reflection image $\mathbf{R}$ captures the intrinsic properties of the object, while the illumination image $\mathbf{L}$ represents the lighting conditions. However, this formulation fails to account for the noise and artifacts introduced by uneven light distribution or dark scenes in low-light conditions. Such artifacts are further exacerbated during image enhancement. To address these issues, we introduce perturbation terms for the illumination and reflection components, denoted as $\tilde{\mathbf{L}}$ and $\tilde{\mathbf{R}}$ respectively, in the original equation: \begin{equation} \mathbf{I} = (\mathbf{R} + \tilde{\mathbf{R}}) \odot (\mathbf{L} + \tilde{\mathbf{L}}), \end{equation} which expands to:
\vspace{-4pt}
\begin{equation} \mathbf{I} = \mathbf{R} \odot \mathbf{L} + \mathbf{R} \odot \tilde{\mathbf{L}} + \tilde{\mathbf{R}} \odot \mathbf{L} + \tilde{\mathbf{R}} \odot \tilde{\mathbf{L}}. \end{equation} 
After simplification, the illuminated image $I_{lu}$ can be expressed as: 
\begin{equation}  \boldsymbol{I}_{lu} = \mathbf{I} \odot \tilde{\mathbf{L}} = \mathbf{R} + \mathbf{C}, \end{equation} 
where $\tilde{\mathbf{L}}$ represents the illumination mapping obtained through convolution for feature extraction, and $\mathbf{C} \in \mathbb{R}^{H \times W \times 3}$ denotes the losses associated with the perturbations. Thus, our Dual Light Enhancement Network can be described as: \begin{equation} 
(\boldsymbol{I}_{lu}, \boldsymbol{F}_{lu}) = \text{LCP}(\mathbf{I}, \boldsymbol{L}_{p}), 
\end{equation} 
\vspace{-15pt}
\begin{equation} \text{DBR}(\boldsymbol{I}_{lu}, \boldsymbol{F}_{lu}) = \text{ILB}(\boldsymbol{I}_{lu}, \boldsymbol{F}_{lu}) + \text{SEB}(\boldsymbol{I}_{lu}),
\end{equation}
\vspace{-15pt}
\begin{equation} \mathbf{DLEN} = \boldsymbol{I}_{lu} + \text{DBR}(\boldsymbol{I}_{lu}, \boldsymbol{F}_{lu}), 
\end{equation} 
where LCP refers to the Light Component Predictor, and DBR represents the Dual-Branch Restorer. The ILB and SEB are the two branches of the DBR, representing the Illumination Learning Branch and the Structure Enhancement Branch, respectively. The LCP takes the low-light image $\mathbf{I}$ and the illumination prior $\boldsymbol{L}_{p} \in \mathbb{R}^{H \times W}$ as input, producing the illuminated image $\boldsymbol{I}_{lu} \in \mathbb{R}^{H \times W \times 3}$ and the feature map $\boldsymbol{F}_{lu} \in \mathbb{R}^{H \times W \times C}$. 
The illumination prior $\boldsymbol{L}_{p}$ is computed by taking the average value of each channel in the image, providing a measure of the overall brightness or illumination level. Consequently, $\boldsymbol{L}_{p}$ functions as an illumination prior, guiding the extraction of lighting information for the image. The outputs of LCP are then fed into the Dual-Branch Restorer (DBR) to mitigate the quality loss exacerbated during illumination enhancement. Before $\boldsymbol{F}_{lu}$ enters the Illumination Learning Branch (ILB), it first passes through the Learnable Wavelet Network (LWN), which aims to further restore the quality loss and enhance the feature learning after the wavelet transformation. The two results from LCP are processed separately by the DBR, resulting in the final restored image, which is represented as $\mathbf{DLEN} \in \mathbb{R}^{H \times W \times 3}$.

\subsection{Dual-Branch Restorer}
Figure.\ref{fig6} illustrates the structure of our Dual-Branch Restorer, which consists of two attention-based branches, each with slightly different architectural designs. In this section, we provide a detailed description of these two branches. The restorer enables spatial interactions and learning across different locations under two conditions, effectively enhancing both the illumination and structural components of the image.

\subsubsection{\textbf{Illumination Learning Branch}} 
This branch employs an encoder-decoder structure. The encoder handles the down-sampling process, while the decoder corresponds to up-sampling. Both processes are symmetric and occur in two stages. In the down-sampling path, $\boldsymbol{I}_{lu}$ undergoes a \(3 \times 3\) convolution, followed by a Multi-Illumination Attention Block (MIAB), and then a \(4 \times 4\) convolution (stride=2) to downscale the features. Two additional MIABs are applied, followed by another \(4 \times 4\) convolution (stride=2) to generate hierarchical features. The up-sampling path is symmetrically designed, using a deconvolution layer (\(2 \times 2\), stride=2) to upscale the features. Skip connections are employed to mitigate information loss during down-sampling. The up-sampling branch outputs a residual image $\boldsymbol{I}_{flb}\in \mathbb{R}^{H \times W \times 3}$. The core unit of the Illumination Learning Branch (ILB) is the Multi-Illumination Attention Block (MIAB), as shown in Figure.\ref{fig6}(b1).

\subsubsection{\textbf{Multi-Illumination Attention Block}} 
As shown in Figure.\ref{fig6}(b1), the light-up feature $\boldsymbol{F}_{lu} \in \mathbb{R}^{H \times W \times C}$, estimated by LCP, is fed into each MIAB. For clarity, Figure.\ref{fig6}(b1) shows the attention mechanism at the largest scale, while smaller scales use \(4 \times 4\) convolutions (stride=2) to downscale $\boldsymbol{F}_{lu}$, which is omitted for simplicity. This block treats a single-channel feature map as a token and computes attention.

The input feature \(F_{in} \in \mathbb{R}^{H \times W \times C}\) is reshaped into tokens \(X \in \mathbb{R}^{HW \times C}\). Then, \(X\) is split into \(k\) heads:
\vspace{-7pt}
\begin{equation}
    X = [X_1, X_2, \cdots, X_k],
\end{equation}
where \(X_i \in \mathbb{R}^{HW \times d_k}\) and \(d_k = \frac{C}{k}\), with \(i = 1, 2, \cdots, k\). Note that Figure.\ref{fig6}(b1) illustrates the case for \(k=1\), omitting some details for simplicity. Each head undergoes a linear projection using fully connected layers without bias to produce query elements \(Q_i\), key elements \(K_i\), and value elements \(V_i\):
\begin{equation}
    Q_i = X_i W_{Q_i}^T, \quad K_i = X_i W_{K_i}^T, \quad V_i = X_i W_{V_i}^T,
\end{equation}
where \(W_{Q_i}, W_{K_i}, W_{V_i} \in \mathbb{R}^{d_k \times d_k}\) are the learnable parameters of the fully connected layers, and \(T\) denotes the matrix transpose. 
Different regions within the same image often exhibit varying lighting conditions. The darker regions are typically associated with more pronounced distortions, making them more challenging to restore. In contrast, areas with better lighting conditions provide richer semantic context, which can facilitate the enhancement of darker regions. To leverage this, we introduce the light-up feature \(F_{lu}\), which encodes lighting information and captures the interactions between regions with disparate lighting conditions. This feature is then used to guide the attention computation. In order to align with the shape of \(X\), we reshape $\boldsymbol{F}_{lu}$ into \(Y \in \mathbb{R}^{HW \times C}\) and decompose it into \(k\) attention heads:
\vspace{-4pt}
\begin{equation}
    Y = [Y_1, Y_2, \cdots, Y_k],
\end{equation}
where \(Y_i \in \mathbb{R}^{HW \times d_k}\) for \(i = 1, 2, \cdots, k\). The self-attention for each head \(i\) is formulated as:
\begin{equation}
    \text{Attention}(Q_i, K_i, V_i, Y_i) = (Y_i \odot V_i) \, \operatorname{softmax}\left(\frac{K_i^T Q_i}{\alpha_i}\right),
\end{equation}
where \(\alpha_i \in \mathbb{R}\) is a learnable parameter that adaptively scales the matrix multiplication. After concatenating the \(k\) heads, the result passes through a fully connected layer and is added to a positional encoding \(P \in \mathbb{R}^{HW \times C}\) (learnable parameters) to produce the output tokens \(X_{out} \in \mathbb{R}^{HW \times C}\). Finally, \(X_{out}\) is reshaped to obtain the output feature \(F_{out} \in \mathbb{R}^{H \times W \times C}\).

\subsubsection{\textbf{Structure Enhancement Branch}} 
The input to the Structure Enhancement branch (Figure.\ref{fig6}(b2)) is also $\boldsymbol{I}_{lu}$. We first apply a convolution to extract low-level feature embeddings. 
The shallow features \(F_0\) are processed through a 4-level symmetric encoder-decoder framework, yielding deep features \(F_d \in \mathbb{R}^{H \times W \times 2C}\). Each encoder-decoder level incorporates multiple Structure Enhancement Attention Blocks (SEAB), with the number of blocks progressively increasing from top to bottom to maintain computational efficiency. The encoder initially operates on the high-resolution input, systematically reducing the spatial dimensions while augmenting the channel capacity. Conversely, the decoder takes the low-resolution latent features \(F_l \in \mathbb{R}^{\frac{H}{8} \times \frac{W}{8} \times 8C}\) and iteratively reconstructs high-resolution representations. Skip connections concatenate encoder features with decoder features, and a \(1 \times 1\) convolution reduces the channel dimension (by half) at all levels except the top one. The SEAB aggregates low-level image features from the encoder with high-level features from the decoder. In the refinement stage, the deep features \(F_{de}\) are further enhanced at high spatial resolution. Finally, a convolution layer is applied to the refined features to generate the restored image \(I_{feb} \in \mathbb{R}^{H \times W \times 3}\). The final restored image is obtained by adding the degraded image to \(I_{feb}\).

The enhanced image $\boldsymbol{I}_{en}$ is derived by summing $\boldsymbol{I}_{lu}$, $\boldsymbol{I}_{flb}$, and $\boldsymbol{I}_{feb}$:
\vspace{-4pt}
\begin{equation}
   \boldsymbol{I}_{en} = \boldsymbol{I}_{lu} + \boldsymbol{I}_{flb} + \boldsymbol{I}_{feb}
\end{equation}
\vspace{-5pt}
\subsubsection{\textbf{Structure Enhancement Attention Block}} 
The Structure Enhancement Attention Block (SEAB) operates on a layer-normalized tensor \( T \in \mathbb{R}^{H \times W \times C} \). The block begins with the generation of query (\(Q\)), key (\(K\)), and value (\(V\)) projections, which are enriched with local contextual information. This is accomplished through a two-step convolutional approach: first, \(1 \times 1\) convolutions are employed to capture pixel-wise cross-channel context, followed by \(3 \times 3\) depth-wise convolutions to encode spatial context at the channel level:
\begin{equation}
    Q_{\mathcal{F}} = W_{\mathcal{D}}^{Q} W_{\mathcal{P}}^{Q} T, \quad K_{\mathcal{F}} = W_{\mathcal{D}}^{K} W_{\mathcal{P}}^{K} T, \quad V_{\mathcal{F}} = W_{\mathcal{D}}^{V} W_{\mathcal{P}}^{V} T,
\end{equation}
where \( W_{\mathcal{P}}^{(\cdot)} \) denotes the \( 1 \times 1 \) point-wise convolution, and \( W_{\mathcal{D}}^{(\cdot)} \) represents the \( 3 \times 3 \) depth-wise convolution. The self-attention process is then defined as:
\vspace{-4pt}
\begin{equation}
\hat{T}_{\mathcal{F}} = W_{\mathcal{P}} \, \text{Attention}\left(\hat{Q}_{\mathcal{F}}, \hat{K}_{\mathcal{F}}, \hat{V}_{\mathcal{F}}\right) + T,
\end{equation}
\begin{equation}
\text{Attention}\left(\hat{Q}_{\mathcal{F}}, \hat{K}_{\mathcal{F}}, \hat{V}_{\mathcal{F}}\right) = \hat{V}_{\mathcal{F}} \cdot \text{Softmax}\left(\frac{\hat{K}_{\mathcal{F}}^T \hat{Q}_{\mathcal{F}}}{\beta_{\mathcal{F}}}\right),
\end{equation}
where \( T \) and \( \hat{T}_{\mathcal{F}} \) represent the input and output feature maps, respectively, and \( \beta_{\mathcal{F}} \) is a learnable scaling parameter that controls the magnitude of the dot product between \( \hat{K}_{\mathcal{F}} \) and \( \hat{Q}_{\mathcal{F}} \) before the softmax operation. After performing element-wise multiplication with the illumination map \( L_{\mathcal{I}} \), the resulting feature map undergoes two additional \( 1 \times 1 \) convolutions, followed by \( 3 \times 3 \) depth-wise convolutions. Finally, it passes through a \( 1 \times 1 \) convolution to output a feature map of the same size as the original input. This operation refines the image features, ensuring that the output preserves both spatial and contextual integrity, ultimately contributing to the enhancement of the restored image.

\subsection{Learnable Wavelet Module}

To more effectively leverage frequency-domain information and preserve image details, we use the Haar wavelet transform, introducing the Learnable Discrete Wavelet Module, which is showed in Figure.\ref{fig6}. By emphasizing high-frequency components, the module significantly improves the reconstruction of local details—such as fine textures and edges—that are often blurred or lost in conventional low-light enhancement methods.

\subsubsection{\textbf{Learnable Discrete Wavelet Transform for Low-Light Image Enhancement}}

For a 1D discrete signal \( g(u) \), the wavelet function is defined as
\vspace{-3pt}
\begin{equation}
    \psi_{m,n}(u) = 2^{m/2} \psi(2^m u - n),
\end{equation}
where \( m \) is the scaling factor, \( n \) is the translation factor, and the scaling function is defined as
\vspace{-3pt}
\begin{equation}
    \phi_{m,n}(u) = 2^{m/2} \phi(2^m u - n).
\end{equation}
The signal decomposition is expressed as:
\begin{equation}
g(u) = \sum_{m > m_0} \sum_n e_{m,n} \psi_{m,n}(u) + \sum_n b_{m_0,n} \phi_{m_0,n}(u), 
\end{equation}
where \( e_{m,n} = \langle g(u),  \psi_{m,n}(u) \rangle \) represents the detail coefficients (high-frequency components), and \( b_{m_0,n} = \langle g(u), \phi_{m_0,n}(u) \rangle \) represents the approximation coefficients (low-frequency components).

\begin{figure*}[ht]
\centering
\includegraphics[width=\textwidth]{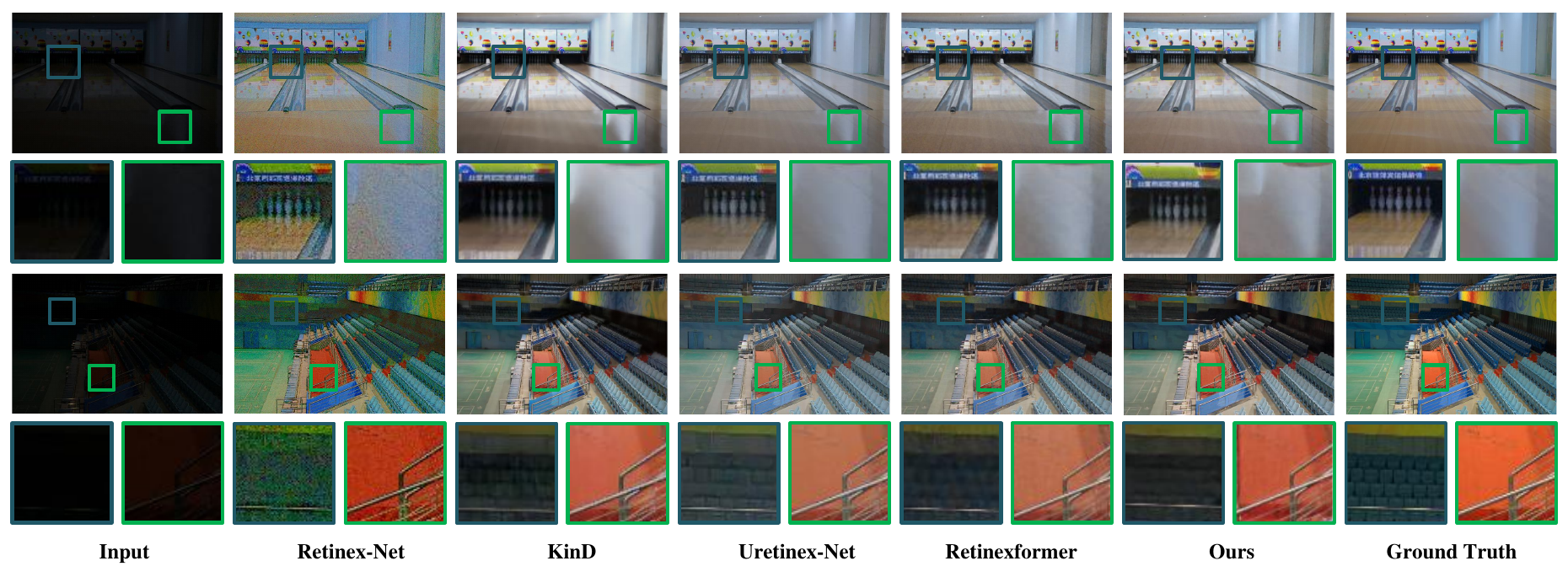} 
\caption{The figure shows the qualitative experimental results on LOLv1. Our method effectively reduces color distortions and enhances lighting effects.}
\label{fig3}
\end{figure*}

To recursively compute the wavelet coefficients, we apply high-pass and low-pass filters, denoted as:
\begin{equation}
e_{m+1,q} = \sum_n h_1[n - 2q] b_{m,n},\quad b_{m+1,q} = \sum_n h_0[n - 2q] b_{m,n},
\end{equation}
where \( h_0 \) and \( h_1 \) are the low-pass and high-pass filters, respectively.

In the 2D case, the wavelet transform decomposes an image into four subbands: low-frequency (\( G_{ll} \)), horizontal high-frequency (\( G_{lh} \)), vertical high-frequency (\( G_{hl} \)), and diagonal high-frequency (\( G_{hh} \)). These subbands are obtained by taking tensor products of the learnable 1D filters \( \vec{h}_0 \) and \( \vec{h}_1 \), as follows:
\begin{align}
G_{ll} &= \vec{h}_0 \otimes \vec{h}_0^T, & G_{lh} &= \vec{h}_0 \otimes \vec{h}_1^T, \\
G_{hl} &= \vec{h}_1 \otimes \vec{h}_0^T, & G_{hh} &= \vec{h}_1 \otimes \vec{h}_1^T, \\
K_g &= \text{cat}(G_{ll}, G_{lh}, G_{hl}, G_{hh}). 
\end{align}
Here, \( \otimes \) represents the outer product, and \( \text{cat} \) denotes the concatenation of the four subbands along the channel dimension. The resulting tensor \( K_g \) captures a multi-resolution representation of the image, which is then used for adaptive feature extraction.

\subsubsection{\textbf{Advantages of Learnable Wavelet for Low-Light Enhancement}}

The use of a learnable wavelet transform in the Light Component Predictor offers several key advantages for low-light image enhancement. First, it allows the network to separate fine high-frequency details, such as edges and textures, from the low-frequency illumination components of the image. This separation ensures that the network can effectively preserve important image details while simultaneously improving overall image quality. Unlike traditional methods that rely on fixed, predefined wavelet bases, our method utilizes learnable wavelet filters, enabling adaptive feature extraction tailored to the low-light enhancement task. This flexibility allows the model to better handle the challenges posed by low-light conditions, such as noise and blurred details, resulting in enhanced image sharpness and improved structure recovery.

\section{Experiments}
\subsection{Datasets and Implementation details}
We evaluated the performance of our model on standard benchmark datasets, specifically the LOL datasets \cite{ref19} \cite{ref20}. The LOL dataset consists of two versions: v1 and v2. The LOLv1 dataset includes 485 training pairs and 15 testing pairs, comprising a total of 500 image pairs, each containing both low-light and normal-light images. The majority of the images focus on interior scenes, and all images have a resolution of 400×600 pixels. The LOLv2 dataset is further divided into two subsets: LOLv2 Real and LOLv2 Synthetic. The training-to-test data ratio for LOLv2 Real is 689:100, and for LOLv2 Synthetic, it is 900:100. The distribution of data pairs and image sizes in LOLv2 is identical to those in LOLv1.
\begin{figure}[ht]
\centering
\includegraphics[width=0.8\columnwidth]{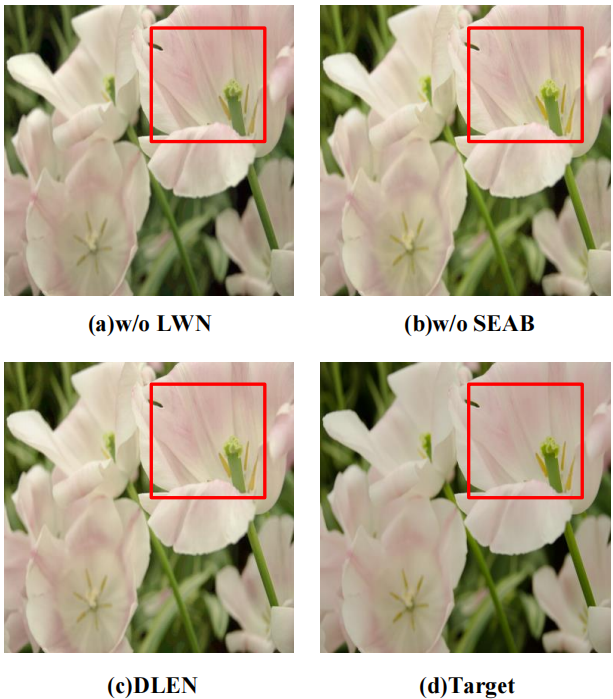} 
\caption{This figure demonstrates the detailed effects of the module on the image. When the LWN module (a) is removed, there is a noticeable loss of texture preservation. Similarly, excluding the SEAB branch (b) results in a loss of crucial structural information. Our method (c), however, produces results that are visually closest to the target (d), with the most accurate preservation of both texture and structural details.}
\vspace{-15pt}
\label{fig5}
\end{figure}
Our method was implemented in PyTorch and experiments were conducted on A10 and A800 GPUs running on a Linux system. We set the image resolution to 128×128 pixels. The batch sizes were set to 8 for LOLv1 and LOLv2 Synthetic, and 4 for LOLv2 Real. Standard data augmentation techniques, including random rotation and flipping, were applied to enhance the training dataset. To minimize the loss, we employed the Adam optimizer. Training objective is to minimize the mean absolute error (MAE) between the enhanced image and ground truth.

\subsection{Evaluation Metrics}
To evaluate the performance of our low-light image enhancement method, we employed two widely used image quality metrics: Peak Signal-to-Noise Ratio (PSNR) and Structural Similarity Index (SSIM). These metrics are commonly used in image enhancement tasks to assess noise levels and structural fidelity, respectively. A higher PSNR indicates superior image enhancement, while a higher SSIM signifies better preservation of high-frequency details and structural integrity.

The Peak Signal-to-Noise Ratio (PSNR) is defined as:
\vspace{-5pt}
\[
\text{PSNR} = 10 \cdot \log_{10} \left( \frac{R^2}{\text{MSE}} \right)
\]

where \( R \) is the maximum possible pixel value of the image (usually 255 for 8-bit images), and MSE (Mean Squared Error) is given by:
\vspace{-5pt}
\[
\text{MSE} = \frac{1}{N \cdot M} \sum_{i=1}^{N} \sum_{j=1}^{M} \left( I(i, j) - K(i, j) \right)^2
\]

where \( I(i, j) \) and \( K(i, j) \) represent the pixel values of the enhanced image and the ground truth image, respectively, and \( N \) and \( M \) are the dimensions of the image.

The Structural Similarity Index (SSIM) is calculated as:
\vspace{-5pt}
\[
\text{SSIM}(x, y) = \frac{(2 \mu_x \mu_y + C_1) (2 \sigma_{xy} + C_2)}{(\mu_x^2 + \mu_y^2 + C_1) (\sigma_x^2 + \sigma_y^2 + C_2)}
\]

where \( \mu_x \) and \( \mu_y \) are the mean intensities of the images \( x \) and \( y \), \( \sigma_x^2 \) and \( \sigma_y^2 \) are the variances of the images, and \( \sigma_{xy} \) is the covariance between the images. \( C_1 \) and \( C_2 \) are small constants used to stabilize the division with weak denominator values, usually set as:
\vspace{-5pt}
\[
C_1 = (k_1 L)^2, \quad C_2 = (k_2 L)^2
\]

where \( L \) is the dynamic range of the pixel values (e.g., 255 for 8-bit images), and \( k_1 \) and \( k_2 \) are constants with typical values \( k_1 = 0.01 \) and \( k_2 = 0.03 \).

\vspace{-5pt}
\subsection{Comparison with other approaches}
\subsubsection{Quantitative Comparisons}

In this work, we present a comprehensive evaluation of our method in comparison to several state-of-the-art (SOTA) techniques across both supervised and unsupervised paradigms, as detailed in Table \ref{tab:mytable1}. The datasets employed for the comparative analysis include both synthetic and real-world data. For clarity, we define "supervised methods" as those trained with reference images, while "unsupervised methods" do not rely on reference images during training. Our method is compared against leading deep learning models, including LIME \cite{ref32}, MBLLEN \cite{ref8}, Retinex-Net \cite{ref30}, KinD \cite{ref31}, KinD++ \cite{ref19}, MIRNet \cite{ref34}, URetinex-Net \cite{ref40}, and Retinexformer \cite{ref11}. The results demonstrate that our DLEN model achieves substantial improvements in Peak Signal-to-Noise Ratio (PSNR), with average increases of 6.734, 6.142, 6.287, 3.878, 4.333, 0.838, 1.584, and 0.505 dB across the two benchmark datasets shown in Table \ref{tab:mytable1}. Additionally, the SSIM values of our method slightly surpass those of other techniques.

\begin{figure*}[ht]
\centering
\includegraphics[width=\textwidth]{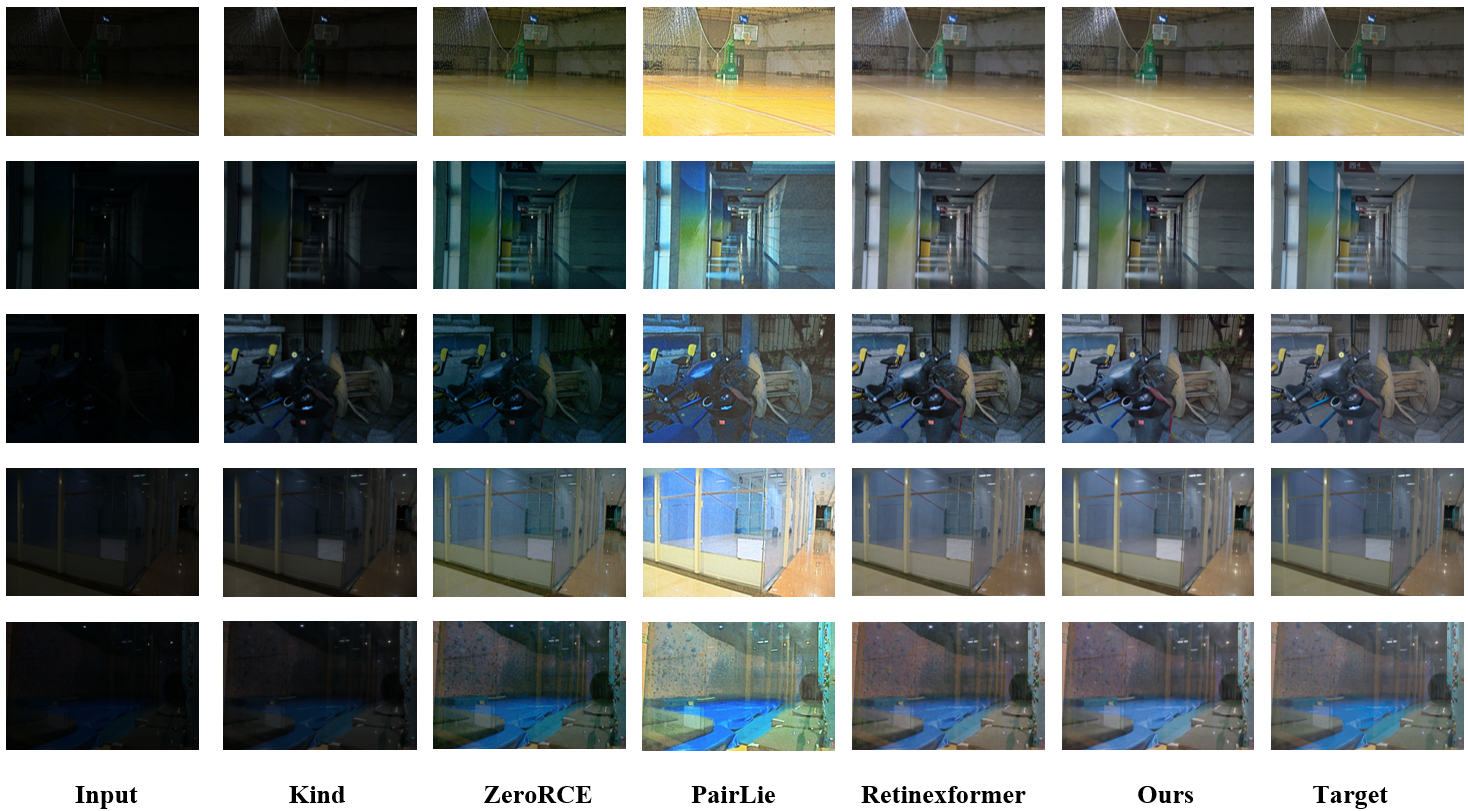} 
\caption{The above shows the qualitative experimental results on LOLv2-real. Our method effectively reduces color distortions and enhances lighting effects.}
\label{fig4}
\vspace{-5pt}
\end{figure*}
\subsubsection{Qualitative Comparisons}


We performed an extensive set of qualitative experiments to visually assess the performance of our method in comparison to other state-of-the-art (SOTA) algorithms. As shown in Figure. \ref{fig1}, the color space recovery results demonstrate that our method closely approximates the ground truth images, highlighting the efficacy of our approach. Figure. \ref{fig2} presents several enhanced sample images, and upon detailed inspection, it is evident that our enhanced images are virtually indistinguishable from the ground truth, indicating a high level of fidelity in preserving both texture and structural details.

A more detailed qualitative comparison between our method, DLEN, and other SOTA algorithms is presented in Figures. \ref{fig3} and \ref{fig4}. Figure. \ref{fig3} compares the results on the LOLv1 dataset, while Figure. \ref{fig4} demonstrates performance on the LOLv2-real dataset. The results from previous methods reveal several notable deficiencies. For example, Retinex-Net \cite{ref30} is prone to noise amplification, while KinD \cite{ref31} tends to generate images with considerable underexposure. PairLie \cite{ref49} exhibits overexposure in certain image regions, and ZeroRCE \cite{ref35} suffers from pronounced noise and artifacts. In addition, Retinexformer \cite{ref11} shows visible underexposure around the bowling ball in Figure. \ref{fig3} and noticeable color distortion in the stadium area at the bottom of the image. In contrast, our DLEN model effectively addresses issues such as exposure management, color distortion, and noise reduction, yielding visually superior enhancement results.

\vspace{-4pt}
\subsection{Ablation Study}
We conduct ablation studies on three datasets: LOLv1, LOLv2-real, and LOLv2-syn. The term "w/o LWN" indicates the removal of the learnable wavelet network in the Illumination Estimator, while "w/o SEAB" refers to the removal of the SEAB branch, retaining only the core architecture. Compared to all other ablation configurations, our full setup achieves the highest PSNR and SSIM values. The "w/o LWN" configuration highlights the limitations of insufficient feature representation in the frequency domain, while "w/o SEAB", which uses only the basic Transformer architecture, suffers from a lack of enhanced structural information.

The visual results of our ablation study are shown in Figure. \ref{fig5}. Direct observation of Figure. \ref{fig5} clearly reveals that the removal of the LWN module leads to a noticeable loss of texture preservation. Similarly, excluding the SEAB branch results in the loss of crucial structural information. In contrast, our method produces results that are visually closest to the target, effectively preserving both texture and structural details.

\begin{table}[ht]
\centering
\caption{Quantitative comparisons on LOL-v1 and LOLv2-real}
\label{tab:mytable1}
\setlength{\tabcolsep}{7pt} 
\begin{tabular}{c|c c|c c}
\hline
\multirow{2}{*}{\textbf{Methods}} & \multicolumn{2}{c|}{\textbf{LOL-v1}} & \multicolumn{2}{c}{\textbf{LOLv2-real}} \\
 & \text{PSNR} $\uparrow$ & \text{SSIM} $\uparrow$  & \text{PSNR} $\uparrow$ & \text{SSIM} $\uparrow$  \\
\hline
\textbf{Supervised} & & & & \\
LIME\cite{ref32} & 16.362 & 0.624  & 16.342 & 0.653  \\
MBLLEN\cite{ref8} & 17.938 & 0.699  & 15.950 & 0.701  \\
Retinex-Net\cite{ref30} & 17.188 & 0.589  & 16.410 & 0.640  \\
KinD\cite{ref31} & 20.347 & 0.813  & 18.070 & 0.781  \\
KinD++\cite{ref19} & 20.707& 0.791  & 16.800 & 0.741  \\
MIRNet \cite{ref34} & \textcolor{red}{24.140} & 0.840  & 20.357 & 0.782  \\
URetinex-Net\cite{ref40} & 21.450 & 0.795 & 21.554 & 0.801  \\
Retinexformer\cite{ref11} & 23.932 & 0.831   & 21.230 & 0.838   \\
DLEN(ours) & 23.942 & \textcolor{red}{0.841}  & \textcolor{red}{22.230} & \textcolor{red}{0.854}  \\
\hline
\textbf{Unsupervised} & & & & \\
GenerativatePrior \cite{ref51} & 12.552 & 0.410 & 13.041 & 0.552  \\
Zero-RCE \cite{ref35} & 16.760 & 0.560  & 18.059 & 0.580 \\
RUAS \cite{ref39} & 16.401 & 0.503  & 16.873 & 0.513  \\
SCI \cite{ref43} & 14.864 & 0.542 & 15.342 & 0.521 \\
PairLie \cite{ref49} & 19.691 & 0.712 & 19.288 & 0.684 \\
NeRCO \cite{ref48} & 19.701 & 0.771 & 19.234 & 0.671\\
CLIP-LIE \cite{ref50} & 17.207 & 0.589 & 17.057 & 0.589\\
\hline
\end{tabular}
\vspace{-10pt}
\end{table}

\begin{table}[ht]
\centering
\caption{Ablation Study on LOL-v1, LOLv2-real and LOLv2-syn}
\label{tab:mytable2}
\setlength{\tabcolsep}{3pt} 
\begin{tabular}{c|c c|c c |c c}
\hline
\multirow{2}{*}{\textbf{Methods}} & \multicolumn{2}{c|}{\textbf{LOL-v1}} & \multicolumn{2}{c|}{\textbf{LOLv2-real}} & \multicolumn{2}{c}{\textbf{LOLv2-syn}} \\
 & \text{PSNR} $\uparrow$ & \text{SSIM} $\uparrow$  & \text{PSNR} $\uparrow$ & \text{SSIM} $\uparrow$  & \text{PSNR} $\uparrow$ & \text{SSIM} $\uparrow$  \\
\hline
w/o LWN & 23.463 &  0.836 & \textcolor{red}{22.293} & 0.853  & 25.788  & 0.933  \\
w/o SEAB & 22.824 & 0.820 & 21.261 & 0.833 & 24.991  & 0.924 \\
DLEN & \textcolor{red}{23.942} & \textcolor{red}{0.841}  & 22.230  & \textcolor{red}{0.854}  & \textcolor{red}{26.261} & \textcolor{red}{0.937}  \\
\hline
\end{tabular}
\vspace{-10pt}
\end{table}

\vspace{-5pt}
\section{Conclusion}
In this paper, we proposed the Dual Light Enhance Network (DLEN) for low-light image enhancement, which introduces a novel architecture combining two distinct attention mechanisms that operate in both spatial and frequency domains. Through the incorporation of a learnable wavelet transform module and a dual-branch architecture, our model preserves high-frequency details and enhances edge and texture information, producing visually natural and high-quality enhanced images.Extensive experimental results on standard benchmarks demonstrate that DLEN outperforms state-of-the-art methods, achieving superior performance in terms of both quantitative metrics and qualitative visual quality. This work highlights the importance of simultaneously considering both spatial and frequency information for low-light image enhancement and provides a promising direction for future research in this area. Future work could explore the application of DLEN in real-time scenarios and the adaptation of the model to more diverse low-light environments, including extremely challenging cases such as night-time and foggy conditions.


\begin{thebibliography}{00}

\bibitem{ref1}
H.-D. Cheng and X.~Shi, ``A simple and effective histogram equalization approach to image enhancement,'' \emph{Digital signal processing}, vol.~14, no.~2, pp. 158--170, 2004.


\bibitem{ref4}
M.~Abdullah-Al-Wadud, M.~H. Kabir, M.~A.~A. Dewan, and O.~Chae, ``A dynamic histogram equalization for image contrast enhancement,'' \emph{IEEE transactions on consumer electronics}, vol.~53, no.~2, pp. 593--600, 2007.


\bibitem{ref6}
C.~Lee, C.~Lee, and C.-S. Kim, ``Contrast enhancement based on layered difference representation of 2d histograms,'' \emph{IEEE transactions on image processing}, vol.~22, no.~12, pp. 5372--5384, 2013.

\bibitem{ref7}
E.~H. Land and J.~J. McCann, ``Lightness and retinex theory,'' \emph{Josa}, vol.~61, no.~1, pp. 1--11, 1971.

\bibitem{ref8}
F.~Lv, F.~Lu, J.~Wu, and C.~Lim, ``Mbllen: Low-light image/video enhancement using cnns.'' in \emph{BMVC}, vol. 220, no.~1.\hskip 1em plus 0.5em minus 0.4em\relax Northumbria University, 2018, p.~4.



\bibitem{ref11}
Y.~Cai, H.~Bian, J.~Lin, H.~Wang, R.~Timofte, and Y.~Zhang, ``Retinexformer: One-stage retinex-based transformer for low-light image enhancement,'' in \emph{Proceedings of the IEEE/CVF International Conference on Computer Vision}, 2023, pp. 12\,504--12\,513.

\bibitem{ref12}
A.~Vaswani, ``Attention is all you need,'' \emph{Advances in Neural Information Processing Systems}, 2017.


\bibitem{ref14}
Y.~Cai, J.~Lin, H.~Wang, X.~Yuan, H.~Ding, Y.~Zhang, R.~Timofte, and L.~V. Gool, ``Degradation-aware unfolding half-shuffle transformer for spectral compressive imaging,'' \emph{Advances in Neural Information Processing Systems}, vol.~35, pp. 37\,749--37\,761, 2022.


\bibitem{ref16}
S.~W. Zamir, A.~Arora, S.~Khan, M.~Hayat, F.~S. Khan, and M.-H. Yang, ``Restormer: Efficient transformer for high-resolution image restoration,'' in \emph{Proceedings of the IEEE/CVF conference on computer vision and pattern recognition}, 2022, pp. 5728--5739.


\bibitem{ref19}
Y.~Zhang, X.~Guo, J.~Ma, W.~Liu, and J.~Zhang, ``Beyond brightening low-light images,'' \emph{International Journal of Computer Vision}, vol. 129, pp. 1013--1037, 2021.

\bibitem{ref20}
W.~Yang, W.~Wang, H.~Huang, S.~Wang, and J.~Liu, ``Sparse gradient regularized deep retinex network for robust low-light image enhancement,'' \emph{IEEE Transactions on Image Processing}, vol.~30, pp. 2072--2086, 2021.


\bibitem{ref23}
S.~Rahman, M.~M. Rahman, M.~Abdullah-Al-Wadud, G.~D. Al-Quaderi, and M.~Shoyaib, ``An adaptive gamma correction for image enhancement,'' \emph{EURASIP Journal on Image and Video Processing}, vol. 2016, pp. 1--13, 2016.


\bibitem{ref26}
Z.-u. Rahman, D.~J. Jobson, and G.~A. Woodell, ``Retinex processing for automatic image enhancement,'' \emph{Journal of Electronic imaging}, vol.~13, no.~1, pp. 100--110, 2004.

\bibitem{ref27}
S.~Wang, J.~Zheng, H.-M. Hu, and B.~Li, ``Naturalness preserved enhancement algorithm for non-uniform illumination images,'' \emph{IEEE transactions on image processing}, vol.~22, no.~9, pp. 3538--3548, 2013.


\bibitem{ref30}
C.~Wei, W.~Wang, W.~Yang, and J.~Liu, ``Deep retinex decomposition for low-light enhancement,'' \emph{arXiv preprint arXiv:1808.04560}, 2018.

\bibitem{ref31}
Y.~Zhang, J.~Zhang, and X.~Guo, ``Kindling the darkness: A practical low-light image enhancer,'' in \emph{Proceedings of the 27th ACM international conference on multimedia}, 2019, pp. 1632--1640.

\bibitem{ref32}
X.~Guo, Y.~Li, and H.~Ling, ``Lime: Low-light image enhancement via illumination map estimation,'' \emph{IEEE Transactions on image processing}, vol.~26, no.~2, pp. 982--993, 2016.

\bibitem{ref33}
K.~G. Lore, A.~Akintayo, and S.~Sarkar, ``Llnet: A deep autoencoder approach to natural low-light image enhancement,'' \emph{Pattern Recognition}, vol.~61, pp. 650--662, 2017.

\bibitem{ref34}
S.~W. Zamir, A.~Arora, S.~Khan, M.~Hayat, F.~S. Khan, M.-H. Yang, and L.~Shao, ``Learning enriched features for real image restoration and enhancement,'' in \emph{Computer Vision--ECCV 2020: 16th European Conference, Glasgow, UK, August 23--28, 2020, Proceedings, Part XXV 16}.\hskip 1em plus 0.5em minus 0.4em\relax Springer, 2020, pp. 492--511.

\bibitem{ref35}
C.~Guo, C.~Li, J.~Guo, C.~C. Loy, J.~Hou, S.~Kwong, and R.~Cong, ``Zero-reference deep curve estimation for low-light image enhancement,'' in \emph{Proceedings of the IEEE/CVF conference on computer vision and pattern recognition}, 2020, pp. 1780--1789.





\bibitem{ref39}
R.~Liu, L.~Ma, J.~Zhang, X.~Fan, and Z.~Luo, ``Retinex-inspired unrolling with cooperative prior architecture search for low-light image enhancement,'' in \emph{Proceedings of the IEEE/CVF conference on computer vision and pattern recognition}, 2021, pp. 10\,561--10\,570.

\bibitem{ref40}
W.~Wu, J.~Weng, P.~Zhang, X.~Wang, W.~Yang, and J.~Jiang, ``Uretinex-net: Retinex-based deep unfolding network for low-light image enhancement,'' in \emph{Proceedings of the IEEE/CVF conference on computer vision and pattern recognition}, 2022, pp. 5901--5910.

\bibitem{ref41}
X.~Xu, R.~Wang, C.-W. Fu, and J.~Jia, ``Snr-aware low-light image enhancement,'' in \emph{Proceedings of the IEEE/CVF conference on computer vision and pattern recognition}, 2022, pp. 17\,714--17\,724.


\bibitem{ref43}
L.~Ma, T.~Ma, R.~Liu, X.~Fan, and Z.~Luo, ``Toward fast, flexible, and robust low-light image enhancement,'' 2022.



\bibitem{ref48}
S.~Yang, M.~Ding, Y.~Wu, Z.~Li, and J.~Zhang, ``Implicit neural representation for cooperative low-light image enhancement,'' in \emph{Proceedings of the IEEE/CVF International Conference on Computer Vision}, 2023, pp. 12\,918--12\,927.

\bibitem{ref49}
Z.~Fu, Y.~Yang, X.~Tu, Y.~Huang, X.~Ding, and K.-K. Ma, ``Learning a simple low-light image enhancer from paired low-light instances,'' in \emph{Proceedings of the IEEE/CVF conference on computer vision and pattern recognition}, 2023, pp. 22\,252--22\,261.

\bibitem{ref50}
Z.~Liang, C.~Li, S.~Zhou, R.~Feng, and C.~C. Loy, ``Iterative prompt learning for unsupervised backlit image enhancement,'' in \emph{Proceedings of the IEEE/CVF International Conference on Computer Vision}, 2023, pp. 8094--8103.

\bibitem{ref51}
X.~Pan, X.~Zhan, B.~Dai, D.~Lin, C.~C. Loy, and P.~Luo, ``Exploiting deep generative prior for versatile image restoration and manipulation,'' \emph{IEEE Transactions on Pattern Analysis and Machine Intelligence}, vol.~44, no.~11, pp. 7474--7489, 2021.


\bibitem{ref53}
Y.~Zhang, W.~Ding, Z.~Pan, and J.~Qin, ``Improved wavelet threshold for image de-noising,'' \emph{Frontiers in neuroscience}, vol.~13, p.~39, 2019.

\bibitem{ref54}
Q.~Fang, Q.~Li, Q.~Song, S.~Montresor, P.~Picart, and H.~Xia, ``Convolutional and fourier neural networks for speckle denoising of wrapped phase in digital holographic interferometry,'' \emph{Optics Communications}, vol. 550, p. 129955, 2024.

\bibitem{ref55}
X.~Gao, T.~Qiu, X.~Zhang, H.~Bai, K.~Liu, X.~Huang, H.~Wei, G.~Zhang, and H.~Liu, ``Efficient multi-scale network with learnable discrete wavelet transform for blind motion deblurring,'' in \emph{Proceedings of the IEEE/CVF Conference on Computer Vision and Pattern Recognition}, 2024, pp. 2733--2742.

\bibitem{ref56}
C.~O. Ancuti and C.~Ancuti, ``Single image dehazing by multi-scale fusion,'' \emph{IEEE Transactions on Image Processing}, vol.~22, no.~8, pp. 3271--3282, 2013.

\bibitem{ref57}
K.~Jiang, Z.~Wang, P.~Yi, C.~Chen, B.~Huang, Y.~Luo, J.~Ma, and J.~Jiang, ``Multi-scale progressive fusion network for single image deraining,'' in \emph{Proceedings of the IEEE/CVF conference on computer vision and pattern recognition}, 2020, pp. 8346--8355.

\bibitem{ref58}
L.~Xiao, C.~Li, Z.~Wu, and T.~Wang, ``An enhancement method for x-ray image via fuzzy noise removal and homomorphic filtering,'' \emph{Neurocomputing}, vol. 195, pp. 56--64, 2016.

\bibitem{ref59}
R.~Fries and J.~Modestino, ``Image enhancement by stochastic homomorphic filtering,'' \emph{IEEE Transactions on Acoustics, Speech, and Signal Processing}, vol.~27, no.~6, pp. 625--637, 1979.

\bibitem{ref60}
C.~Dong, C.~C. Loy, K.~He, and X.~Tang, ``Image super-resolution using deep convolutional networks,'' \emph{IEEE transactions on pattern analysis and machine intelligence}, vol.~38, no.~2, pp. 295--307, 2015.

\bibitem{ref61}
J.~Bai, Y.~Yin, Q.~He, Y.~Li, and X.~Zhang, ``Retinexmamba: Retinex-based mamba for low-light image enhancement,'' \emph{arXiv preprint arXiv:2405.03349}, 2024.

\bibitem{ref62}
M.~Li, H.~Sun, Y.~Lei, X.~Zhang, Y.~Dong, Y.~Zhou, Z.~Li, and X.~Chen, ``High-fidelity document stain removal via a large-scale real-world dataset and a memory-augmented transformer,'' in \emph{Proc. WACV}, 2024.

\bibitem{ref63}
X.~Guo, X.~Chen, S.~Wang, and C.-M.~Pun, ``Underwater image restoration through a prior guided hybrid sense approach and extensive benchmark analysis,'' \emph{IEEE TCSVT}, 2025.

\bibitem{ref64}
Z.~Li, X.~Chen, C.-M.~Pun, and X.~Cun, ``High-resolution document shadow removal via a large-scale real-world dataset and a frequency-aware shadow erasing net,'' in \emph{Proc. ICCV}, 2023, pp.~12449--12458.

\bibitem{ref65}
X.~Guo, X.~Chen, S.~Luo, S.~Wang, and C.-M.~Pun, ``Dual-hybrid attention network for specular highlight removal,'' in \emph{Proc. ACM MM}, 2024, pp.~10173--10181.

\bibitem{ref66}
X.~Guo, Y.~Dong, X.~Chen, W.~Chen, Z.~Li, F.~Zheng, and C.-M.~Pun, ``Underwater image restoration via polymorphic large kernel CNNs,'' in \emph{Proc. ICASSP}, 2025, pp.~1--5.

\bibitem{ref67}
Z.~Zhou, Y.~Lei, X.~Chen, S.~Luo, W.~Zhang, C.-M.~Pun, and Z.~Wang, ``Docdeshadower: Frequency-aware transformer for document shadow removal,'' in \emph{Proc. SMC}, 2024, pp.~2468--2473.

\bibitem{ref68}
F.~Zheng, X.~Chen, W.~Liu, H.~Li, Y.~Lei, J.~He, C.-M.~Pun, and S.~Zhou, ``SMAFormer: Synergistic multi-attention transformer for medical image segmentation,'' in \emph{Proc. BIBM}, 2024, pp.~4048--4053.

\bibitem{ref69}
X.~Chen, B.~Lei, C.-M.~Pun, and S.~Wang, ``Brain diffuser: An end-to-end brain image to brain network pipeline,'' in \emph{Proc. PRCV}, 2023, pp.~16--26.

\bibitem{ref70}
S.~Luo, X.~Chen, W.~Chen, Z.~Li, S.~Wang, and C.-M.~Pun, ``Devignet: High-resolution vignetting removal via a dual aggregated fusion transformer with adaptive channel expansion,'' in \emph{Proc. AAAI}, 2024, pp.~4000--4008.

\bibitem{ref71}
Z.~Li, X.~Chen, S.~Wang, and C.-M.~Pun, ``A large-scale film style dataset for learning multi-frequency driven film enhancement,'' in \emph{Proc. IJCAI}, 2023, pp.~1160--1168.

\bibitem{ref72}
X.~Chen, C.-M.~Pun, and S.~Wang, ``Medprompt: Cross-modal prompting for multi-task medical image translation,'' in \emph{Proc. PRCV}, 2024, pp.~61--75.

\bibitem{ref73}
L.~Zhu, W.~Liu, X.~Chen, Z.~Li, X.~Chen, Z.~Wang, and C.-M.~Pun, ``Test-time intensity consistency adaptation for shadow detection,'' \emph{arXiv}, 2024.

\bibitem{ref74}
Y.~Huo, G.~Huang, L.~Cheng, J.~He, X.~Chen, X.~Yuan, G.~Zhong, and C.-M.~Pun, ``IMAN: An adaptive network for robust NPC mortality prediction with missing modalities,'' in \emph{Proc. BIBM}, 2024, pp.~2074--2079.

\bibitem{ref75}
D.~Zhu, G.~Huang, X.~Yuan, X.~Chen, G.~Zhong, C.-M.~Pun, and J.~Deng, ``FAQNet: Frequency-aware quaternion network for endoscopic highlight removal,'' in \emph{Proc. BIBM}, 2024, pp.~1408--1413.

\bibitem{ref76}
X.~Zhang, F.~Chen, C.~Wang, M.~Tao, and G.-P.~Jiang, ``Sienet: Siamese expansion network for image extrapolation,'' \emph{IEEE Signal Processing Letters}, vol.~27, pp.~1590--1594, 2020.

\bibitem{ref77}
X.~Zhang, Y.~Zhao, C.~Gu, C.~Lu, and S.~Zhu, ``SpA-Former: An effective and lightweight transformer for image shadow removal,'' in \emph{Proc. IJCNN}, 2023, pp.~1--8.

\bibitem{ref78}
Z.~Xu, X.~Zhang, W.~Chen, J.~Liu, T.~Xu, and Z.~Wang, ``MuralDiff: Diffusion for ancient murals restoration on large-scale pre-training,'' \emph{IEEE Transactions on Emerging Topics in Computational Intelligence}, 2024.













\end{thebibliography}
\end{document}